\documentclass[letterpaper, 10 pt, conference]{ieeeconf}  

\IEEEoverridecommandlockouts                              

\usepackage{float}
\usepackage{graphics} 
\usepackage{epsfig} 
\usepackage{mathptmx} 
\usepackage{times} 
\usepackage{amsmath} 
\usepackage{amssymb}  
\usepackage{colortbl}
\usepackage[T1]{fontenc}
\usepackage{wrapfig,lipsum}
\usepackage{booktabs}
\usepackage{bigstrut}
\usepackage{multirow}
\usepackage{placeins}
\usepackage{gensymb}
\usepackage{xcolor}
\usepackage{soul}
\usepackage[ruled,linesnumbered, noend]{algorithm2e}
\usepackage{tikz}
\usepackage{yfonts}
\usepackage{graphicx}
\usepackage{import}
\usepackage{newtxtext}
\usepackage{newtxmath}
\usepackage{hhline}
\usepackage{stfloats}
\usepackage{hyperref}
\usepackage{cite}
\usepackage{setspace}
\usepackage{atbegshi,picture}
\usepackage[font={scriptsize}]{caption}

\newenvironment{myitem}{\begin{list}{$\bullet$}
{\setlength{\itemsep}{-0pt}
\setlength{\topsep}{0pt}
\setlength{\labelwidth}{0pt}
\setlength{\leftmargin}{10pt}
\setlength{\parsep}{-0pt}
\setlength{\itemsep}{0pt}
\setlength{\partopsep}{0pt}}}%
{\end{list}}

\hypersetup{
pdfauthor={Bowen Wen, Wenzhao Lian, Kostas Bekris and Stefan Schaal},
pdftitle={CaTGrasp: Learning Category-Level Task-Relevant Grasping in Clutter from Simulation},
pdfsubject={robotics; computer vision; artificial intelligence},
pdfkeywords={grasping; pose estimation; perception},
}

\AtBeginShipoutNext{\AtBeginShipoutUpperLeft{%
  \put(\dimexpr\paperwidth-0.4cm\relax,-0.6cm){\makebox[0pt][r]{\framebox{\scriptsize Accepted in IEEE International Conference on Robotics and Automation (ICRA) 2022}}}%
}}

\title{\LARGE \bf CaTGrasp: Learning Category-Level Task-Relevant\\ Grasping in Clutter from Simulation}

\author{Bowen Wen$^{1, 2}$, Wenzhao Lian$^{1}$, Kostas Bekris$^{2}$ and Stefan Schaal$^{1}$
\thanks{\scriptsize $^{1}$Intrinsic Innovation LLC in CA, USA. {\{wenzhaol, sschaal\}@intrinsic.ai}. This research was conducted during Bowen's internship at Intrinsic. }
\thanks{\scriptsize
$^{2}$Rutgers University in NJ, USA.
\{bw344, kostas.bekris\}@cs.rutgers.edu. Bowen Wen and Kostas Bekris were partially supported by the US NSF Grant IIS-1734492. The opinions expressed here are of the authors and do not reflect the views of the sponsor.}
}

\begin{document}

\maketitle
\thispagestyle{empty}
\pagestyle{empty}

\begin{abstract}
Task-relevant grasping is critical for industrial assembly, where downstream manipulation tasks constrain the set of valid grasps. Learning how to perform this task, however, is challenging, since task-relevant grasp labels are hard to define and annotate. There is also yet no consensus on proper representations for modeling or off-the-shelf tools for performing task-relevant grasps.  This work proposes a framework to learn task-relevant grasping for industrial objects without the need of time-consuming real-world data collection or manual annotation. To achieve this, the entire framework is trained solely in simulation, including supervised training with synthetic label generation and self-supervised, hand-object interaction. In the context of this framework, this paper proposes a novel, object-centric canonical representation at the category level, which allows establishing dense correspondence across object instances and transferring task-relevant grasps to novel instances. Extensive experiments on task-relevant grasping of densely-cluttered industrial objects are conducted in both simulation and real-world setups, demonstrating the effectiveness of the proposed framework. Code and data are available at \url{https://sites.google.com/view/catgrasp}.
\end{abstract}

\section{Introduction}

Robot manipulation often requires identifying a suitable grasp that is aligned with a downstream task. An important application domain is industrial assembly, where the robot needs to perform constrained placement after grasping an object \cite{andrewbwrss,luo2021robust}. In such cases, a suitable grasp requires stability during object grasping and transporting while avoiding obstructing the placement process. For instance, a grasp where the gripper fingers cover the thread portion of a screw can impede its placement through a hole. Grasping a screw in this manner is not a task-relevant grasp.

There are many challenges, however, in solving task-relevant grasps. (a) Grasping success and task outcome are mutually dependent \cite{montesano2008learning}. (b) Task-relevant grasping involves high-level semantic information, which cannot be easily modeled or represented. (c) 6D grasp pose annotation in 3D is more complicated than 2D image alternatives. Achieving task-relevant grasps requires additional semantic priors in the label generation process than geometric grasp generation often studied in stable grasp learning \cite{mahler2017dex,ten2017grasp}. (d) Generalization to highly-variant, novel instances within the category requires effective learning on category-level priors. (e) In densely cluttered industrial object picking scenarios, as considered in this work, the growing number of objects, the size of the grasp solution space, as well as challenging object properties (e.g., textureless, reflective, tiny objects, etc.), introduce additional combinatorial challenges and demand increased robustness.

With recent advances in deep learning, many efforts resort to semantic part segmentation \cite{chu2019learning,monica2020point} or keypoint detection \cite{manuelli2019kpam} via supervised learning on manually labeled real-world data. A line of research is circumventing the complicated data collection process by training in simulation \cite{fang2020learning,qin2020keto}. It still remains an open question, however, whether category-level priors can be effectively captured through end-to-end training. Moreover, most of the existing work considers picking and manipulation of isolated objects \cite{wang2021efficient,song2015task,mavrakis2016task,gaobuffer2021,gao2021fast}. The complexity of densely cluttered industrial object picking scenarios considered in this work, make the task-relevant grasping task even more challenging.

\begin{figure}[t]
  \centering
  \includegraphics[width=0.48\textwidth]{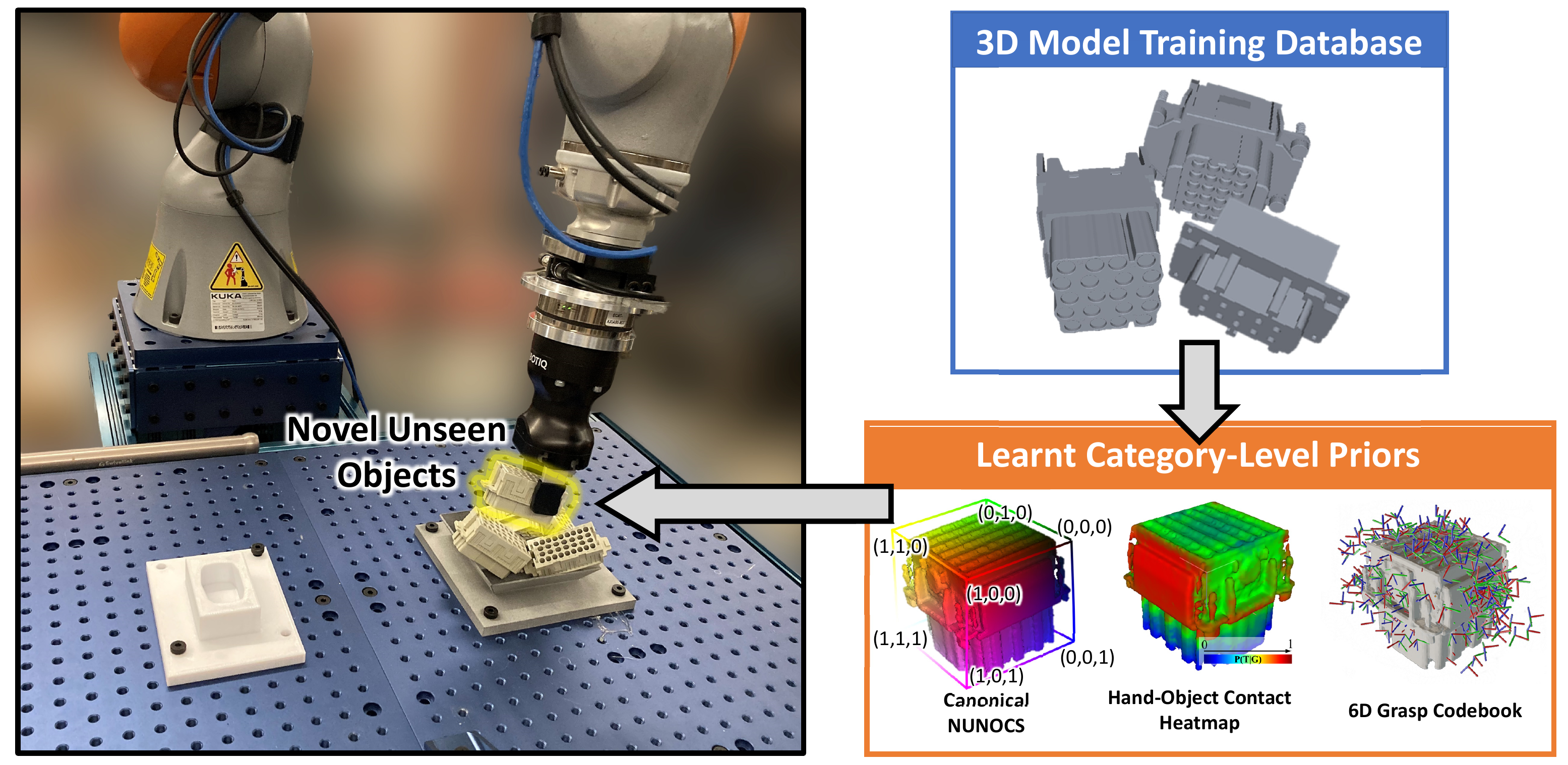}
  \vspace{-0.25in}
  \caption{\scriptsize Given a database of 3D models of same category, the proposed method learns: (a) an object-centric NUNOCS representation that is canonical for the object category, (b) a heatmap that indicates the task achievement success likelihood dependent on the hand-object contact region during the grasp, and (c) a codebook of stable 6D grasp poses. The heatmap and the grasp poses are transferred to real-world, novel unseen object instances during testing for solving task-relevant grasping.}
  \label{fig:intro}
  \vspace{-0.25in}
\end{figure}

To tackle the above challenges, this work aims to learn category-level, task-relevant grasping solely in simulation, circumventing the requirement of manual data collection or annotation efforts. In addition, during the test stage, the trained model can be directly applied to novel object instances with previously unseen dimensions and shape variations, saving the effort of acquiring 3D models or re-training for each individual instance. This is achieved by encoding 3D properties - including shape, pose and gripper-object contact experience that is relevant to task performance - shared across diverse instances within the category. Therefore, once trained, this category-level, task-relevant grasping knowledge not only transfers across novel instances, but also effectively generalizes to real-world densely cluttered scenarios without the need for fine-tuning. In summary, the contributions of this work are the following:

\begin{myitem}

\item A novel framework for learning category-level, task-relevant grasping of densely cluttered industrial objects and targeted placement. To the best of the authors' knowledge, this is the first work that effectively tackles task-relevant grasping of industrial objects in densely cluttered scenarios in a scalable manner without human annotation.

\item Instead of learning sparse keypoints relevant to task-relevant manipulation, as in \cite{qin2020keto, manuelli2019kpam}, this work models dense, point-wise task relevance on 3D shapes. To do so, it leverages hand-object contact heatmaps generated in a self-supervised manner in simulation. This dense 3D representation eliminates the requirement of manually specifying keypoints.

\item It introduces the "Non-Uniform Normalized Object Coordinate Space" (NUNOCS) representation for learning category-level object 6D poses and 3D scaling, which allows non-uniform scaling across three dimensions. Compared to the previously proposed Normalized Object Coordinate Space (NOCS) representation \cite{wang2019normalized} developed in the computer vision community for category-level 6D pose and 1D uniform scale estimation, the proposed representation allows to establish more reliable dense correspondence and thus enables fine-grained knowledge transfer across object instances with large shape variations.

\item The proposed framework is solely trained in simulation and generalizes to the real-world without any re-training, by leveraging domain randomization \cite{tobin2017domain}, bi-directional alignment \cite{wense3tracknet}, and domain-invariant, hand-object contact heatmaps modeled in a category-level canonical space. To this end, a synthetic training data generation pipeline, together with its produced novel dataset in industrial dense clutter scenarios, is presented.

\end{myitem}


\section{Related Work}
\vspace{-0.05in}

\noindent \textbf{Stable Grasping} - Stable grasping methods focus on robust grasps. The methods can be generally classified into two categories: model-based and model-free. Model-based grasping methods require object CAD models to be available beforehand for computing and storing grasp poses offline w.r.t. specific object instances. During test stage, the object's 6D pose is estimated to transform the offline trained grasp poses to the object in the scene\cite{zeng2017multi,xiang2017posecnn,mitash2020scene,wen2020robust}. 
More recently, model-free methods relax this assumption by directly operating over observations such as raw point cloud \cite{mahler2017dex, mahler2019learning} or images \cite{park2020real,cheng2020high,baichuanvisual2022,huangdipn2021,huang2022selfsupervised}, or transferring the category-level offline trained grasps via Coherent Point Drift \cite{rodriguez2018transferring}. Representative works \cite{mahler2017dex,ten2017grasp,liang2019pointnetgpd} train a grasping evaluation network to score and rank the grasp candidates sampled over the observed point cloud. For the sake of efficiency, more recent works \cite{mousavian20196,qin2020s4g,fang2020graspnet,breyer2020volumetric,park2020real,sundermeyer2021contact} develop grasp pose prediction networks, which directly output 6D grasp proposals along with their scores, given the scene observation. 
Differently, the proposed \texttt{CaTGrasp} aims to compute grasps that are not only stable but also task-relevant. 


\noindent \textbf{Task-Relevant Grasping} - Task-relevant grasping requires the grasps to be compatible with downstream manipulation tasks. Prior works have developed frameworks to predict affordance segmentation \cite{do2018affordancenet,detry2017task,monica2020point,antanas2019semantic,kokic2017affordance,ardon2020self} or keypoints \cite{xu2021affordance} over the observed image or point cloud. This, however, often assumes manually annotated real world data is available to perform supervised training \cite{kokic2020learning,allevato2020learning,song2015task}, which is costly and time-consuming to obtain. While \cite{chu2019learning,yang2019task} alleviates the problem via sim-to-real transfer, it still requires manual specification of semantic parts on 3D models for generating synthetic affordance labels. Instead, another line of research \cite{fang2020learning, qin2020keto,turpin2021gift} proposed to learn semantic tool manipulation via self-interaction in simulation. While the above research commonly tackles the scenarios of tool manipulation or household objects, \cite{zhao2020towards} shares the closest setting to ours in terms of industrial objects. In contrast to the above, our work considers more challenging densely cluttered scenarios. It also generalizes to novel unseen object instances, without requiring objects' CAD models for pose estimation or synthetic rendering during testing as in \cite{zhao2020towards}. 


\noindent \textbf{Category-Level Manipulation} - In order to generalize to novel objects without CAD models, category-level manipulation is often achieved by learning correspondence shared among similar object instances, via dense pixel-wise representation \cite{florence2018dense,yang2021learning,chai2019multi} or semantic keypoints \cite{manuelli2019kpam,qin2020keto}. 
In particular, sparse keypoint representations are often assigned priors about their semantic functionality and require human annotation.
While promising results on household objects and tools have been shown\cite{manuelli2019kpam,qin2020keto}, it becomes non-trivial to manually specify semantic keypoints for many industrial objects (Fig. \ref{fig:realworld_bar}), where task-relevant grasp poses can be widely distributed. Along the line of work on manipulation with dense correspondence, \cite{florence2018dense} developed a framework for learning dense correspondence over 2D image pairs by training on real world data, which is circumvented in our case. \cite{yang2021learning,chai2019multi} extended this idea to multi-object manipulation given a goal configuration image. Instead of reasoning on 2D image pairs which is constrained to specific view points, this work proposes NUNOCS representation to establish dense correspondence in 3D space. This direct operation in 3D allows to transfer object-centric contact experience, along with a 6D grasp pose codebook. Additionally, the task-relevant grasping has not been achieved in \cite{florence2018dense,yang2021learning,chai2019multi}.



\begin{figure*}[h]
  \centering
  \definecolor{red}{RGB}{255,0,0}
  \vspace{+0.05in}
  \includegraphics[width=0.98\textwidth]{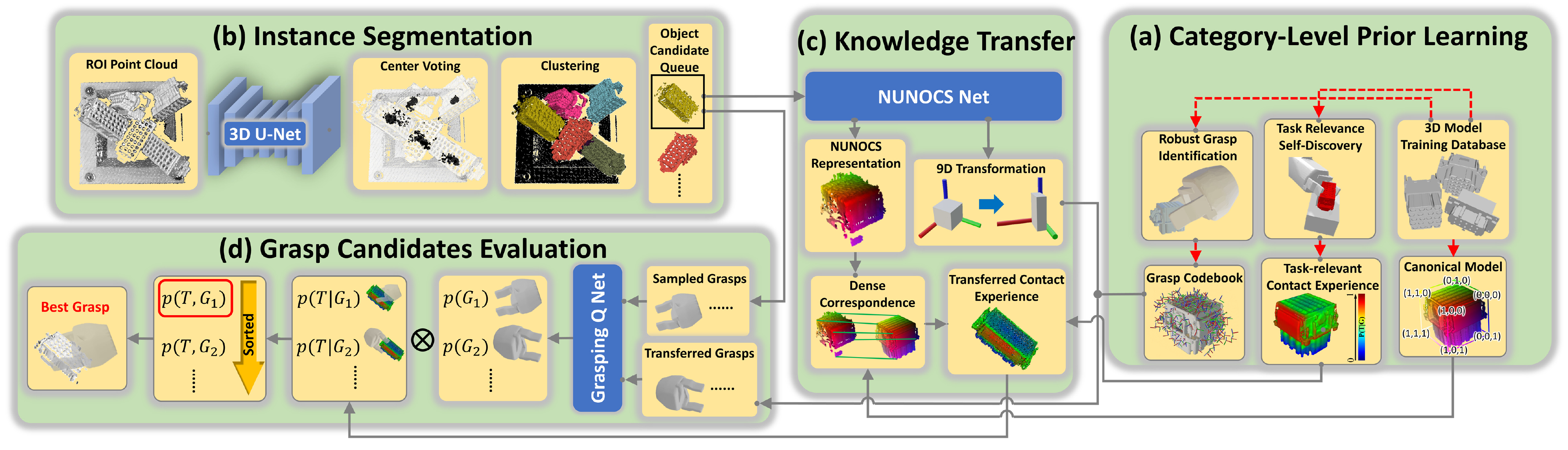}
  \vspace{-0.1in}
  \caption{\scriptsize Overview of the proposed framework. \textbf{Right: (a)}  Given a collection of CAD models for objects of the same category, the NUNOCS representation is aggregated to generate a canonical model for the category. The CAD models are further utilized in simulation to generate synthetic point cloud data for training all the networks (\textit{3D U-Net, NUNOCS Net} and \textit{Grasping Q Net}). Meanwhile, the category-level grasp codebook and hand-object contact heatmap are identified via self-interaction in simulation. \textbf{Top-left: (b)} A \textit{3D U-Net} is leveraged to predict point-wise centers of objects in dense clutter, based on which the instance segmentation is computed by clustering. \textbf{Center: (c)}  The \textit{NUNOCS Net} operates over an object's segmented point cloud and predicts its NUNOCS representation to establish dense correspondence with the canonical model and compute its 9D pose $\xi_o \in  \{SE(3)\times R^{3}\}$ (6D pose and 3D scaling). This allows to transfer the precomputed category-level knowledge to the observed scene. \textbf{Bottom-left: (d)}  Grasp proposals are generated both by transferring them from a canonical grasp codebook and directly by sampling over the observed point cloud. IK-infeasible or in-collision (using FCL \cite{pan2012fcl}) grasps are rejected. Then, the \textit{Grasping Q Net} evaluates the stability of the accepted grasp proposals. This information is combined with a task-relevance score computed from the grasp's contact region. The entire process can be repeated for multiple object segments to find the currently best grasp to execute according to $P(T,G)=P(T|G)P(G)$. Red dashed arrows occur in the offline training stage only.}
  \label{fig:pipeline}
  \vspace{-0.25in}
\end{figure*}

\vspace{-0.05in}
\section{Problem Statement}
\vspace{-0.05in}

We assume novel unseen objects of the same type have been collected into a bin, forming a densely cluttered pile as in common industrial settings. The objective is to compute task-relevant 6D grasp poses $\xi_{G} \in SE(3)$ that allow a downstream constrained placement task. The grasping process is repeated for each object instance until the bin is cleared. The inputs to the framework are listed below.
\begin{myitem}
\item A collection of 3D models $\mathcal{M}_{C}$ belonging to category $C$ for training (e.g., Fig \ref{fig:realworld_bar} left). This does not include any testing instance in the same category, i.e., $M^{\text{test}}_{C} \notin \mathcal{M}_{C}$.
\item A downstream placement task $T_C$ corresponding to the category (e.g., Fig \ref{fig:qual_grasp}), including a matching receptacle and the criteria of placement success.
\item A depth image $I_{D}$ of the scene for grasp planning at test stage.
\end{myitem}

\section{Approach}
\vspace{-0.05in}
Fig. \ref{fig:pipeline} summarizes the proposed framework. Offline, given a collection of models $\mathcal{M}_{C}$ of the same category, synthetic data are generated in simulation (Sec. \ref{sec:sim_data}) for training the \textit{NUNOCS Net} (Sec. \ref{sec:nunocs}), \textit{Grasping Q Net} (Sec. \ref{sec:stable_grasp}) and \textit{3D U-Net} (Sec. \ref{sec:instance_seg}). Then, self-interaction in simulation provides hand-object contact experience, which is summarized in task-relevant heatmaps for grasping (Sec. \ref{sec:afford}). The canonical NUNOCS representation allows the aggregation of category-level, task-relevant knowledge across instances. Online, the category-level knowledge is transferred from the canonical NUNOCS model to the segmented target object via dense correspondence and 9D pose estimation, guiding the grasp candidate generation and selection.

\vspace{-0.05in}
\subsection{Category-level Canonical NUNOCS representation}
\label{sec:nunocs}
\vspace{-0.05in}
Previous work \cite{florence2018dense} learned dense correspondence between object instances using contrastive loss. It requires training on real-world data and operates over 2D images from specific viewpoints. Instead, this work establishes dense correspondence in 3D space to transfer knowledge from a trained model database $\mathcal{M}_C$ to a novel instance $M_{C}^{\text{test}}$.  

Inspired by \cite{wang2019normalized}, this work presents the Non-Uniform Normalized Object Coordinate Space (NUNOCS) representation. Given an instance model $M$, all the points are normalized along each dimension, to reside within a unit cube:
\begin{gather*} 
p^{d}_{\mathbb{C}}=(p^{d}-p^{d}_{min})/(p^{d}_{max}-p^{d}_{min}) \in [0,1]; d\in \{x,y,z\}.
\end{gather*} 
The transformed points exist in the canonical NUNOCS $\mathbb{C}$ (Fig. \ref{fig:intro} bottom-right). In addition to being used for synthetic training data generation (Sec. \ref{sec:sim_data}), the models $\mathcal{M}_C$ are also used to create a category-level canonical template model, to generate a hand-object contact heatmap (Sec. \ref{sec:afford}) and a stable grasp codebook (Sec. \ref{sec:stable_grasp}). To do so, each model in $\mathcal{M}_C$ is converted to the space $\mathbb{C}$, and the canonical template model is represented by the one with the minimum sum of Chamfer distances to all other models in $\mathcal{M}_C$. The transformation from each model to this template is then utilized for aggregating the stable grasp codebook and the task-relevant hand-object contact heatmap.

For the \textit{NUNOCS Net}, we aim to learn $\Phi: \mathcal{P}_{o} \rightarrow \mathcal{P}_{\mathbb{C}}$, where $\mathcal{P}_{o}$ and $\mathcal{P}_{\mathbb{C}}$ are the observed object cloud and the canonical space cloud, respectively. $\Phi(\cdot)$ is built with a PointNet-like architecture \cite{qi2017pointnet} given it is light-weight and efficient. The learning task is formulated as a classification problem by discretizing $p^{d}_{\mathbb{C}}$ into 100 bins. Softmax cross entropy loss is used as we found it more effective than regression by reducing the solution space \cite{wang2019normalized}. Along with the predicted dense correspondence, the 9D object pose $\xi_{o}\in \{SE(3) \times R^{3}\}$ is also recovered. It is computed via \textit{RANSAC}\cite{fischler1981random} to provide an affine transformation from the predicted canonical space cloud $\mathcal{P}_{\mathbb{C}}$ to the observed object segment cloud $\mathcal{P}_{o}$, while ensuring the rotation component to be orthonormal.

\setlength{\columnsep}{0.05in}%
\setlength{\intextsep}{0.in}%
\begin{wrapfigure}{r}{0.2\textwidth}
  \centering
  \includegraphics[width=0.2\textwidth]{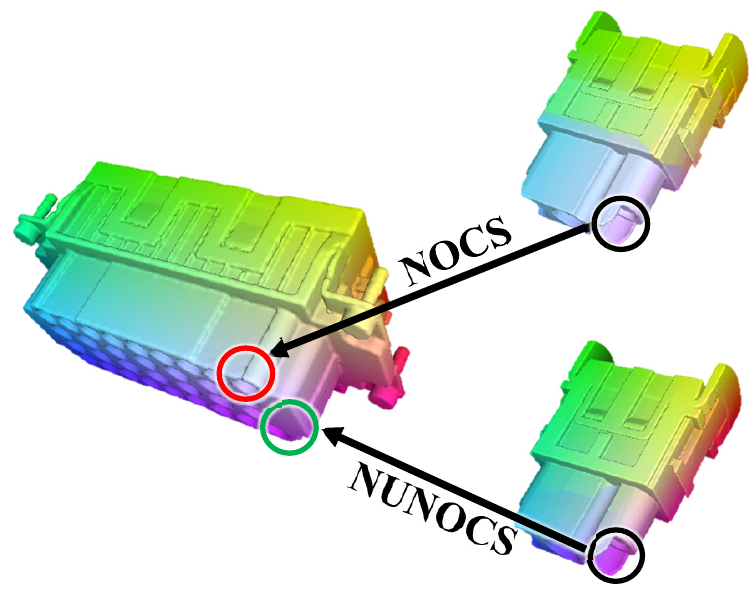} 
\end{wrapfigure}

Compared to the original NOCS representation \cite{wang2019normalized}, which recovers a 7D pose of novel object instances, the proposed NUNOCS allows to scale independently in each dimension when converting to the canonical space. Therefore, more fine-grained dense correspondence across object instances can be established via measuring their similarity ($L_{2}$ distance in our case) in $\mathbb{C}$. This is especially the case for instances with dramatically different 3D scales, as shown in the wrapped figure, where colors indicate dense correspondence similarity in $\mathbb{C}$ and one example correspondence for NUNOCS and NOCS respectively. A key difference from another related work on VD-NOC \cite{patten2020dgcm}, which directly normalizes the scanned point cloud in the camera frame, is that the proposed NUNOCS representation is object-centric and thus agnostic to specific camera parameters or viewpoints.

\vspace{-0.05in}
\subsection{Stable Grasp Learning}
\label{sec:stable_grasp}
\vspace{-0.05in}
During offline training, grasp poses are uniformly sampled from the point cloud of each object instance, covering the feasible grasp space around the object. For each grasp $G$, the grasp quality is evaluated in simulation. To compute a continuous score $s_{G} \in [0,1]$ as training labels, 50 neighboring grasp poses are randomly sampled in the proximity of $\xi_{G}\in SE(3)$ and executed to compute the empirical grasp success rate. The intuition is that grasp stability should be continuous over its 6D neighborhood. Once the grasps are generated, they are then exploited in two ways. 


First, given the relative 9D transformation from the current instance to the canonical model, the grasp poses are converted into the NUNOCS space and stored in a \textit{stable grasp codebook} $\mathcal{G}$. During test time, given the estimated 9D object pose $\xi_o \in \{SE(3)\times R^{3}\}$ of the observed object's segment relative to the canonical space $\mathbb{C}$, grasp proposals can be generated by applying the same transformation to the grasps in $\mathcal{G}$. Compared with traditional online grasp sampling over the raw point cloud \cite{ten2017grasp,mahler2017dex}, this grasp knowledge transfer is also able to generate grasps from occluded object regions. In practice, the two strategies can be combined to form a robust hybrid mode for grasp proposal generation.

Second, the generated grasps are utilized for training the \textit{Grasping Q Net}, which is built based on PointNet \cite{qi2017pointnet}. Specifically, in each dense clutter generated (as in Sec. \ref{sec:sim_data}), the object segment in the 3D point cloud is transformed to the grasp's local frame given the object and grasp pose. The \textit{Grasping Q Net} takes the point cloud as input and predicts the grasp's quality $P(G)$, which is then compared against the discretized grasp score $s_{G}$ to compute softmax cross entropy loss. This one-hot score representation has been observed to be effective for training \cite{liang2019pointnetgpd}. 

\subsection{Affordance Self-Discovery}\label{sec:afford}
\begin{figure}[h]
  \centering
  \vspace{-0.1in}
  \includegraphics[width=0.48\textwidth]{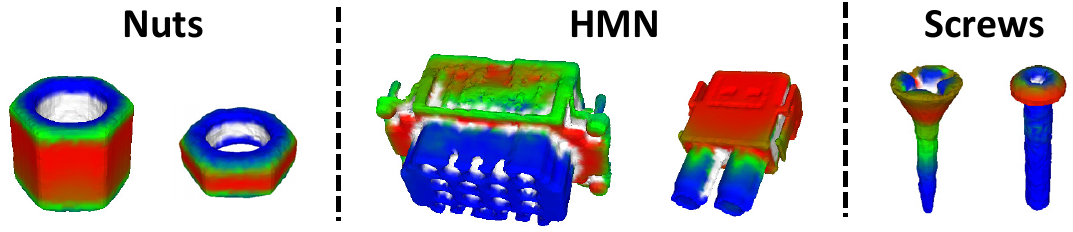}
  \vspace{-0.2in}
  \caption{\scriptsize Examples of task-relevant hand-object contact heatmaps $P(T|G)$. Warmer color indicates higher values of $P(T|G)$. The white areas are the small concave regions for which the rigid parallel-jaw gripper can't touch. They remain unexplored and set to the default $P(T|G)=0.5$ though they are also unlikely to be touched during testing. The collected contact heatmap is object-centric and domain-invariant. Once identified in simulation, it is directly applied in the real world.}
  \label{fig:qual_heatmap}
  \vspace{-0.1in}
\end{figure}

\vspace{-0.05in}
In contrast to prior work \cite{monica2020point}, which manually annotates parts of segments, or uses predefined sparse keypoints \cite{manuelli2019kpam}, this work discovers grasp affordance via self-interaction. In particular, the objective is to compute $P(T|G)=P(T,G)/P(G)$ automatically for all graspable regions on the object. To achieve this, a dense 3D point-wise hand-object contact heatmap is modeled. For each grasp in the codebook $G\in \mathcal{G}$ (generated as in Sec. \ref{sec:stable_grasp}), a grasping process is first simulated. The hand-object contact points are identified by computing their signed distance w.r.t the gripper mesh. If it's a stable grasp, i.e., the object is lifted successfully against gravity, the count $n(G)$ for all contacted points on the object are increased by $1$. Otherwise, the grasp is skipped. For these stable grasps, a placement process is simulated, i.e., placing the grasped object on a receptacle, to verify the task relevance (Fig. \ref{fig:pipeline} top-right). Collision is checked between the gripper and the receptacle during this process. If the gripper does not obstruct the placement and if the object can steadily rest in the receptacle, the count of joint grasp and task success $n(G,T)$ on the contact points is increased by $1$. After all grasps are verified, for each point on the object point cloud, its task relevance can be computed as $P(T|G)=n(G,T)/n(G)$. Examples of self-discovered hand-object contact heatmaps are shown in Fig. \ref{fig:qual_heatmap}. Interestingly, these heatmaps achieve similar performance to human annotated part-segmentation \cite{do2018affordancenet} but can be interpreted as a ``soft'' version. 

Eventually, for each of the training objects within the category, the hand-object contact heatmap $P(T|G)$ is transformed to the canonical model. The task-relevant heatmaps over all training instances are aggregated and averaged to be the final canonical model's task-relevance heatmap. During testing, due to the partial view of the object's segment, the antipodal contact points $p_{c}$ are identified between the gripper mesh and the transformed canonical model (Fig. \ref{fig:pipeline} bottom-left). For each grasp candidate, the score $P_G(T|G)=\frac{1}{|p_c|}\sum_{p_c} P_{p_c}(T|G)$ is computed. It is then combined with the predicted $P_G(G)$ from \textit{Grasping Q Net} (Sec. \ref{sec:stable_grasp}) to compute the grasp's task-relevance score: $P_{G}(T,G)=P_G(T|G)P_G(G)$. 

\vspace{0.15in}

\subsection{Instance Segmentation in Dense Clutter}\label{sec:instance_seg}

This work employs the Sparse 3D U-Net \cite{jiang2020pointgroup,SubmanifoldSparseConvNet} due to its memory efficiency. The network takes as input the entire scene point cloud $\mathcal{P} \in R^{N \times 3}$ voxelized into sparse volumes and predicts per point offset $\mathcal{P}_{\text{offset}} \in R^{N \times 3}$ w.r.t. to predicted object centers. The training loss is designed as the $L_{2}$ loss between the predicted and the ground-truth offsets \cite{xie2021unseen}. The network is trained independently, since joint end-to-end training with the following networks has been observed to cause instability during training.

During testing, the predicted offset is applied to the original points, leading the shifted point cloud to condensed point groups $\mathcal{P}+\mathcal{P}_{\text{offset}}$, as shown in Fig. \ref{fig:pipeline}. Next, DBSCAN \cite{ester1996density} is employed to cluster the shifted points into instance segments. Additionally, the segmented point cloud is back-projected onto the depth image $I_D$ to form 2D segments. This provides an approximation of the per-object visibility by counting the number of pixels in each segment. Guided by this, the remaining modules of the framework prioritize the top layer of objects given their highest visibility in the pile during grasp candidate generation. 
\vspace{-0.05in}
\subsection{Training Data Generation in Simulation}
\label{sec:sim_data}
\vspace{-0.05in}

The entire framework is trained solely on synthetic data. To do so, synthetic data are generated in PyBullet simulation \cite{coumans2021}, aiming for physical plausibility \cite{tremblay2018deep,wense3tracknet}, while leveraging domain randomization \cite{tobin2017domain} and bi-directional domain alignment on the depth modality \cite{wense3tracknet}. At the start of each scene generation, an object instance type and its scale is randomly chosen from the associated category's 3D model database $\mathcal{M}_C$. The number of object instances in the bin, the camera pose relative to the bin, and physical parameters (such as bounciness and friction) are randomized. To generate dense clutter, object poses are randomly initialized above the bin. The simulation is executed until the in-bin objects  are stabilized. The ground-truth labels for NUNOCS, grasping quality and instance segmentation are then retrieved from the simulator.


\begin{figure*}[t]
  \centering
  \definecolor{blue}{RGB}{67, 143, 211}
  \definecolor{red}{RGB}{215, 83, 106}
    \vspace{+0.1in}
  \includegraphics[width=0.98\textwidth]{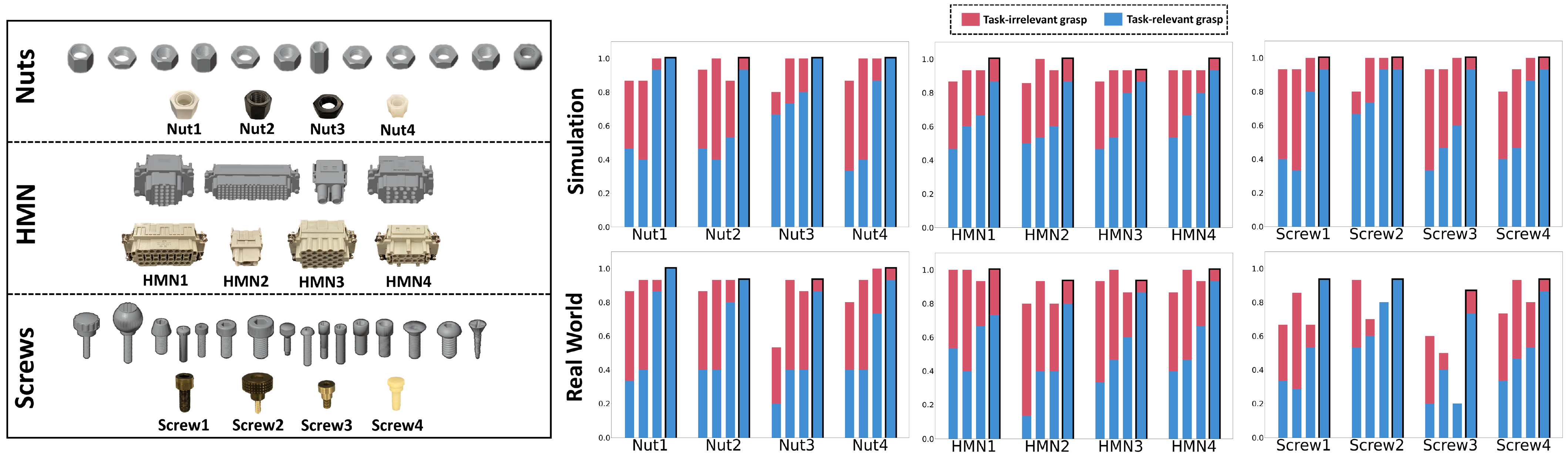}
  \vspace{-0.1in}
  \caption{\scriptsize \textbf{Left:} The 3 object categories: \textit{Nuts}, \textit{HMN} and \textit{Screws}. For each category, the first row is a collection of 3D models used for learning in simulation. The second row is the novel unseen  instances with challenging properties (tiny, texture-less, glossy, etc), used during testing both in simulation and in the real-world. \textbf{Right:} Instance-level grasp performance evaluated in simulation (top) and real-world (bottom). For each object instance, the group of 4 stacked bars from left to right are \texttt{PointNetGPD} \cite{liang2019pointnetgpd}, \texttt{Ours-NA}, \texttt{Ours-NOCS} and \texttt{Ours}, where the column corresponding to \texttt{Ours} is marked with a black boundary. Stable grasps include both \textcolor{red}{task-irrelevant} and \textcolor{blue}{task-relevant} grasps, while the missing blanks are grasp failures.}
  \label{fig:realworld_bar}
  \vspace{-0.3in}
\end{figure*}

\section{Experiments}
\vspace{-0.05in}

This section aims to experimentally evaluate $3$ questions: i) Is the proposed dense correspondence expressive and reliable enough to represent various object instances within a category? ii) How well does the proposed model, only trained in simulation, generalize to real-world settings for task-relevant grasping? iii) Does the proposed category-level knowledge learning also benefit grasp stability? Our proposed method is compared against:

\begin{myitem}
\item {\texttt{PointNetGPD} \cite{liang2019pointnetgpd}}:
A state-of-the-art method on robust grasping. Its open-source code\footnote{\scriptsize \url{https://github.com/lianghongzhuo/PointNetGPD}} is adopted. For fair comparison, the network is retrained using the same synthetic training data of industrial objects as our method. At test time, it directly samples grasp proposals over the raw point cloud without performing instance segmentation \cite{liang2019pointnetgpd}.

\item {\texttt{Ours-NA}}:
A variant of our method that does not consider task-relevant affordance but still transfers category-level grasp knowledge. Only $P(G)$ is used for ranking grasp candidates. 

\item {\texttt{Ours-NOCS}}:
A variant of our method by replacing the NUNOCS representation with NOCS \cite{wang2019normalized} for solving the category-level pose, while the remainings are the same as our framework. This serves as an ablation to study the effectiveness of capturing cross-instance large variations by using the proposed NUNOCS representation. 
\end{myitem}

Some other alternatives are not easy to compare against directly. For instance, \texttt{KETO} \cite{qin2020keto} and \texttt{kPAM} \cite{manuelli2019kpam} focused on singulated household object picking from table-top, which is not easy to adapt to our setting. In addition, \texttt{kPAM} requires human annotated real-world data for training.

\subsection{Experimental Setup}
\vspace{-0.05in}
In this work, 3 different industrial object categories are included: \textit{Nuts}, \textit{Screws} and \textit{HMN} series connectors. Their training and testing splits are depicted in Fig. \ref{fig:realworld_bar}(left). For each category, the first row is a collection of 3D models crawled online. This inexpensive source for 3D model training is used to learn category-level priors. The second row is 4 novel unseen object instances used for testing. Different from the training set, the testing object instances are real industrial objects purchased from retailers\footnote{\scriptsize \url{https://www.digikey.com}; \url{https://www.mcmaster.com}} for realistic evaluation. They are examined and ensured to be novel unseen object instances separated from the training set. The testing object instances are chosen so as to involve sufficient variance to evaluate the cross-instance generalization of the proposed method, while being graspable by the gripper in our configuration. The CAD models of the testing object instances are solely used for setting up the simulation environment.

\setlength{\columnsep}{0.05in}  
\setlength{\intextsep}{0.0in}     
\begin{wrapfigure}{r}{0.3\textwidth}
  \centering
  \includegraphics[width=0.3\textwidth]{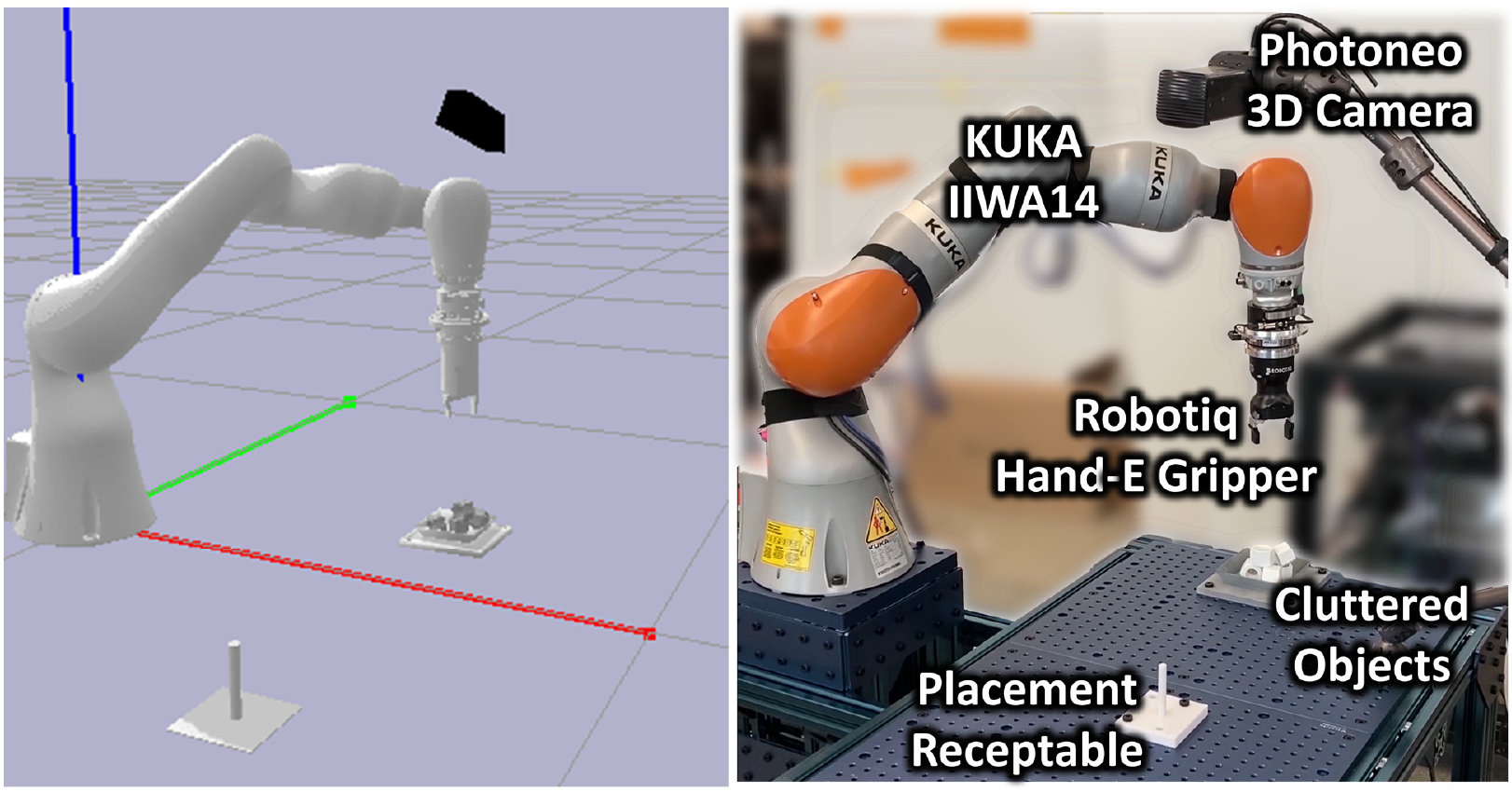} 
\end{wrapfigure}

Evaluations are performed in similar setups in simulation and the real-world. The hardware is composed of a Kuka IIWA14 arm, a Robotiq Hand-E gripper, and a Photoneo 3D camera, as in the wrapped figure. Simulation experiments are conducted in PyBullet, with the corresponding hardware components modeled and gravity applied to manipulated objects. At the start of the bin-picking process, a random number (between 4 to 6) of object instances of the same type are randomly placed inside the bin to form a cluttered pile. Experiments for each of the 12 object instances have been repeated 10 times in simulation and 3 times in real-world, with different arbitrarily formed initial pile configurations. This results in approximately 600 and 180 grasp evaluations in simulation and real-world respectively for each evaluated approach.
For each bin-clearing scenario, its initial pile configuration is recorded and set similarly across all evaluated methods for fair comparison.

After each grasp, its stability is evaluated by a lifting action. If the object drops, the grasp is marked as failure. For stable grasps, additional downstream category-specific placement tasks are performed to further assess the task-relevance. A stable grasp is further examined and marked as a task-relevant grasp, if the placement also succeeds. Otherwise, it is marked as a task-irrelevant grasp, though being stable. The placement receptacles are CAD designed and 3D printed for each object instance with tight placement tolerances ($<3mm$). For evaluation purposes, the placement planning is performed based on manually annotated 6D in-hand object pose post-grasping. This effort is beyond the scope of this work.

\begin{table}[h]
\centering
\resizebox{0.45\textwidth}{!}{
\begin{tabular}{l||ccc|c}
\hline
      & Nuts  & HMN   & Screws & Total \bigstrut\\
\hline
\hline
PointnetGPD & 53.3\% & 49.2\% & 45.0\% & 49.2\% \bigstrut[t]\\
Ours-NA & 51.1\% & 58.3\% & 50.0\% & 53.1\% \\
Ours-NOCS & 75.6\% & 71.7\% & 80.0\% & 75.7\% \\
\rowcolor[rgb]{ .749,  .749,  .749} Ours & \textbf{97.8\%} & \textbf{88.3\%} & \textbf{93.3\%} & \textbf{93.1\%} \bigstrut[b]\\
\hline
\end{tabular}%
}
\vspace{-0.05in}
\caption{\scriptsize  Results of \textbf{task-relevant} grasp percentage out of the total grasp attempts in simulation. For each method, approximately 600 grasps are conducted.}
\label{tab:sim}
\end{table}

\vspace{+0.1in}
\begin{table}[h]
\centering
\resizebox{0.45\textwidth}{!}{
\begin{tabular}{l||ccc|c}
\hline
      & Nuts  & HMN   & Screws & Total \bigstrut\\
\hline
\hline
PointnetGPD & 33.3\% & 35.0\% & 38.0\% & 35.4\% \bigstrut[t]\\
Ours-NA & 40.0\% & 43.3\% & 42.9\% & 42.1\% \\
Ours-NOCS & 70.0\% & 58.3\% & 52.5\% & 60.3\% \\
\rowcolor[rgb]{ .749,  .749,  .749} Ours & \textbf{93.3\%} & \textbf{83.3\%} & \textbf{86.7\%} & \textbf{87.8\%} \bigstrut[b]\\
\hline
\end{tabular}%
}
\vspace{-0.05in}
\caption{\scriptsize Results of \textbf{task-relevant} grasp percentage out of the total grasp attempts in real-world. For each method, approximately 180 grasps are conducted.}
\label{tab:realworld}
\end{table}


\begin{figure*}[t]
  \centering
  \definecolor{green}{RGB}{0, 176, 80}
  \definecolor{blue}{RGB}{0, 176, 240}
  \definecolor{red}{RGB}{255,0,0}
  \vspace{+0.1in}
  \includegraphics[width=0.98\textwidth]{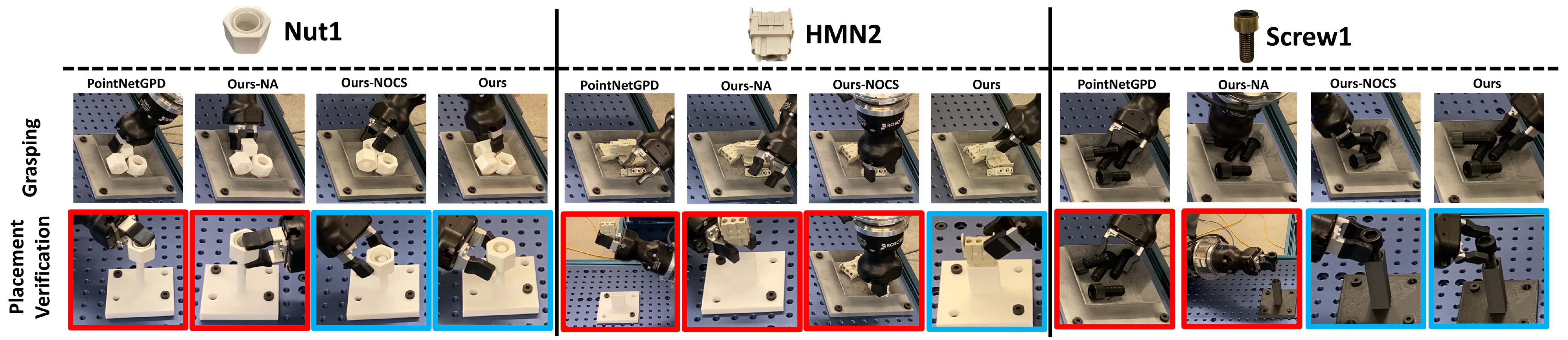}
  \vspace{-0.1in}
  \caption{\scriptsize Qualitative comparison of the grasping and placement evaluation in the real-world. The snapshots are taken for one of the test objects per category during the bin-clearing process, where the initial pile configuration is similar across different methods. For the placement verification images (second row), \textcolor{red}{red} boxes indicate either failure grasps, or stable but task-irrelevant grasps. \textcolor{blue}{Blue} boxes indicate task-relevant grasps resulting in successful placement. Note that given the similar pile configuration, the methods do not necessarily choose the same object instance as the target due to different grasp ranking strategies. See the supplementary media for the complete video.}
  \label{fig:qual_grasp}
  \vspace{-0.3in}
\end{figure*}

\subsection{Results and Analysis}
\vspace{-0.05in}
The quantitative results in simulation and real-world are shown in Table \ref{tab:sim} and \ref{tab:realworld} respectively. The success rate excludes the task-irrelevant or failed grasps.  As demonstrated in the two tables, \texttt{Ours} significantly surpasses all baselines measured by the success rate on task-relevant grasping in both simulation and real-world. Example real-world qualitative results are shown in Fig. \ref{fig:qual_grasp}.

Fig.~\ref{fig:realworld_bar} (right) decomposes the semantic grasping attempts of all the methods at the object instance level. Thanks to the task-relevant hand-object contact heatmap modeling and knowledge transfer via 3D dense correspondence, most of the grasps generated by \texttt{Ours} and \texttt{Ours-NOCS} are task-relevant semantic grasps. In comparison, a significant percentage of grasps planned by \texttt{PointNetGPD} and \texttt{Ours-NA} are not semantically meaningful, i.e., the objects, though stably grasped, cannot be directly manipulated for the downstream task without regrasping.

The fact that \texttt{Ours} reaches comparable or better performance than \texttt{Ours-NOCS} indicates that the proposed NUNOCS is a more expressive representation for 3D dense correspondence modeling and task-relevant grasp knowledge transfer. In particular, for the object ``HMN2'' (Fig. \ref{fig:realworld_bar} left), the performance gap is more noticeable as its 3D scales vary significantly from the canonical model along each of the 3 dimensions. Additionally, the number of 3D training models available for the HMN category is more limited compared to the \textit{Screws} or \textit{Nuts} category. This requires high data-efficiency to capture the large variance. Despite these adverse factors, \texttt{Ours} is able to learn category-level task-relevant knowledge effectively, by virtue of the more representative NUNOCS space.

Although our proposed method targets at task-relevance, it also achieves a high stable grasp success rate, as shown in Fig.~\ref{fig:realworld_bar} (right). This demonstrates the efficacy of the proposed hybrid grasp proposal generation, where additional grasps transferred from the category-level grasp codebook span a more complete space around the object including occluded parts. This is also partially reflected by comparing \texttt{Ours-NA} and \texttt{PointNetGPD}, where \texttt{PointNetGPD} generates grasp candidates by solely sampling over the observation point cloud.

Comparing the overall performance across simulation and real-world experiments indicates a success rate gap of a few percent. This gap is noticeably larger for \textit{Screws}, of which the instances are thin and tiny. Therefore, when they rest in a cluttered bin, it is challenging to find collision-free grasps and thus requires high precision grasp pose reasoning. In particular, as shown in Fig. \ref{fig:realworld_bar}, the instance ``Screw3'' challenges all evaluated methods in terms of grasp stability.
Improving the gripper design \cite{ha2020fit2form} for manipulating such tiny objects is expected to further elevate the performance. In addition, during gripper closing, the physical dynamics of \textit{Screws} is challenging to model when they roll inside the bin in simulation. With more advanced online domain adaptation techniques \cite{antonova2018global}, the performance is expected to be boosted.


\section{Conclusion}
\label{sec:conclusion}
\vspace{-0.05in}
This work proposed \texttt{CaTGrasp} to learn task-relevant grasping for industrial objects in simulation, avoiding the need for time-consuming real-world data collection or manual annotation. With a novel object-centric canonical representation at the category level, dense correspondence across object instances is established and used for transferring task-relevant grasp knowledge to novel object instances. Extensive experiments on task-relevant grasping of densely cluttered industrial objects are conducted in both simulation and real-world setups, demonstrating the method's effectiveness. In future work, developing a complete task-relevant framework with visual tracking \cite{wen2021bundletrack} in the feedback loop for manipulating novel unseen objects is of interest.


\bibliographystyle{IEEEtran}
\bibliography{ref}  

\begin{thebibliography}{10}
\providecommand{\url}[1]{#1}
\csname url@samestyle\endcsname
\providecommand{\newblock}{\relax}
\providecommand{\bibinfo}[2]{#2}
\providecommand{\BIBentrySTDinterwordspacing}{\spaceskip=0pt\relax}
\providecommand{\BIBentryALTinterwordstretchfactor}{4}
\providecommand{\BIBentryALTinterwordspacing}{\spaceskip=\fontdimen2\font plus
\BIBentryALTinterwordstretchfactor\fontdimen3\font minus
  \fontdimen4\font\relax}
\providecommand{\BIBforeignlanguage}[2]{{%
\expandafter\ifx\csname l@#1\endcsname\relax
\typeout{** WARNING: IEEEtran.bst: No hyphenation pattern has been}%
\typeout{** loaded for the language `#1'. Using the pattern for}%
\typeout{** the default language instead.}%
\else
\language=\csname l@#1\endcsname
\fi
#2}}
\providecommand{\BIBdecl}{\relax}
\BIBdecl

\bibitem{andrewbwrss}
A.~S. Morgan, B.~Wen, J.~Liang, A.~Boularias, A.~M. Dollar, and K.~Bekris,
  ``Vision-driven compliant manipulation for reliable, high-precision assembly
  tasks,'' \emph{RSS}, 2021.

\bibitem{luo2021robust}
J.~Luo, O.~Sushkov, R.~Pevceviciute, W.~Lian, C.~Su, M.~Vecerik, N.~Ye,
  S.~Schaal, and J.~Scholz, ``Robust multi-modal policies for industrial
  assembly via reinforcement learning and demonstrations: A large-scale
  study,'' \emph{RSS}, 2021.

\bibitem{montesano2008learning}
L.~Montesano, M.~Lopes, A.~Bernardino, and J.~Santos-Victor, ``Learning object
  affordances: from sensory--motor coordination to imitation,'' \emph{IEEE
  Transactions on Robotics}, vol.~24, no.~1, pp. 15--26, 2008.

\bibitem{mahler2017dex}
J.~Mahler, J.~Liang, S.~Niyaz, M.~Laskey, R.~Doan, X.~Liu, J.~A. Ojea, and
  K.~Goldberg, ``Dex-net 2.0: Deep learning to plan robust grasps with
  synthetic point clouds and analytic grasp metrics,'' \emph{RSS}, 2017.

\bibitem{ten2017grasp}
A.~ten Pas, M.~Gualtieri, K.~Saenko, and R.~Platt, ``Grasp pose detection in
  point clouds,'' \emph{The International Journal of Robotics Research},
  vol.~36, no. 13-14, pp. 1455--1473, 2017.

\bibitem{chu2019learning}
F.-J. Chu, R.~Xu, and P.~A. Vela, ``Learning affordance segmentation for
  real-world robotic manipulation via synthetic images,'' \emph{IEEE Robotics
  and Automation Letters}, vol.~4, no.~2, pp. 1140--1147, 2019.

\bibitem{monica2020point}
R.~Monica and J.~Aleotti, ``Point cloud projective analysis for part-based
  grasp planning,'' \emph{IEEE Robotics and Automation Letters}, vol.~5, no.~3,
  pp. 4695--4702, 2020.

\bibitem{manuelli2019kpam}
L.~Manuelli, W.~Gao, P.~Florence, and R.~Tedrake, ``{KPAM}: Keypoint
  affordances for category-level robotic manipulation,'' \emph{ISRR}, 2019.

\bibitem{fang2020learning}
K.~Fang, Y.~Zhu, A.~Garg, A.~Kurenkov, V.~Mehta, L.~Fei-Fei, and S.~Savarese,
  ``Learning task-oriented grasping for tool manipulation from simulated
  self-supervision,'' \emph{The International Journal of Robotics Research},
  vol.~39, no. 2-3, pp. 202--216, 2020.

\bibitem{qin2020keto}
Z.~Qin, K.~Fang, Y.~Zhu, L.~Fei-Fei, and S.~Savarese, ``{KETO}: Learning
  keypoint representations for tool manipulation,'' in \emph{2020 IEEE
  International Conference on Robotics and Automation (ICRA)}.\hskip 1em plus
  0.5em minus 0.4em\relax IEEE, 2020, pp. 7278--7285.

\bibitem{wang2021efficient}
R.~Wang, Y.~Miao, and K.~E. Bekris, ``Efficient and high-quality prehensile
  rearrangement in cluttered and confined spaces,'' \emph{Arxiv}, 2021.

\bibitem{song2015task}
D.~Song, C.~H. Ek, K.~Huebner, and D.~Kragic, ``Task-based robot grasp planning
  using probabilistic inference,'' \emph{IEEE transactions on robotics},
  vol.~31, no.~3, pp. 546--561, 2015.

\bibitem{mavrakis2016task}
N.~Mavrakis, M.~Kopicki, R.~Stolkin, A.~Leonardis \emph{et~al.},
  ``Task-relevant grasp selection: A joint solution to planning grasps and
  manipulative motion trajectories,'' in \emph{2016 IEEE/RSJ International
  Conference on Intelligent Robots and Systems (IROS)}.\hskip 1em plus 0.5em
  minus 0.4em\relax IEEE, 2016, pp. 907--914.

\bibitem{gaobuffer2021}
\BIBentryALTinterwordspacing
K.~Gao, S.~Feng, and J.~Yu, ``On minimizing the number of running buffers for
  tabletop rearrangement,'' \emph{Robotics: Science and Systems XVII}, Jul
  2021. [Online]. Available: \url{http://dx.doi.org/10.15607/RSS.2021.XVII.033}
\BIBentrySTDinterwordspacing

\bibitem{gao2021fast}
K.~Gao, D.~Lau, B.~Huang, K.~E. Bekris, and J.~Yu, ``Fast high-quality tabletop
  rearrangement in bounded workspace,'' \emph{arXiv}, 2021.

\bibitem{wang2019normalized}
H.~Wang, S.~Sridhar, J.~Huang, J.~Valentin, S.~Song, and L.~J. Guibas,
  ``Normalized object coordinate space for category-level 6d object pose and
  size estimation,'' in \emph{Proceedings of the IEEE/CVF Conference on
  Computer Vision and Pattern Recognition}, 2019, pp. 2642--2651.

\bibitem{tobin2017domain}
J.~Tobin, R.~Fong, A.~Ray, J.~Schneider, W.~Zaremba, and P.~Abbeel, ``Domain
  randomization for transferring deep neural networks from simulation to the
  real world,'' in \emph{IROS 2017}.

\bibitem{wense3tracknet}
\BIBentryALTinterwordspacing
B.~Wen, C.~Mitash, B.~Ren, and K.~E. Bekris, ``se(3)-tracknet: Data-driven 6d
  pose tracking by calibrating image residuals in synthetic domains,''
  \emph{2020 IEEE/RSJ International Conference on Intelligent Robots and
  Systems (IROS)}, Oct 2020. [Online]. Available:
  \url{http://dx.doi.org/10.1109/IROS45743.2020.9341314}
\BIBentrySTDinterwordspacing

\bibitem{zeng2017multi}
A.~Zeng, K.-T. Yu, S.~Song, D.~Suo, E.~Walker, A.~Rodriguez, and J.~Xiao,
  ``Multi-view self-supervised deep learning for 6d pose estimation in the
  amazon picking challenge,'' in \emph{2017 IEEE international conference on
  robotics and automation (ICRA)}.\hskip 1em plus 0.5em minus 0.4em\relax IEEE,
  2017, pp. 1386--1383.

\bibitem{xiang2017posecnn}
Y.~Xiang, T.~Schmidt, V.~Narayanan, and D.~Fox, ``Posecnn: A convolutional
  neural network for 6d object pose estimation in cluttered scenes,''
  \emph{RSS}, 2018.

\bibitem{mitash2020scene}
C.~Mitash, B.~Wen, K.~Bekris, and A.~Boularias, ``Scene-level pose estimation
  for multiple instances of densely packed objects,'' in \emph{Conference on
  Robot Learning}.\hskip 1em plus 0.5em minus 0.4em\relax PMLR, 2020, pp.
  1133--1145.

\bibitem{wen2020robust}
B.~Wen, C.~Mitash, S.~Soorian, A.~Kimmel, A.~Sintov, and K.~E. Bekris,
  ``Robust, occlusion-aware pose estimation for objects grasped by adaptive
  hands,'' in \emph{2020 IEEE International Conference on Robotics and
  Automation (ICRA)}.\hskip 1em plus 0.5em minus 0.4em\relax IEEE, 2020, pp.
  6210--6217.

\bibitem{mahler2019learning}
J.~Mahler, M.~Matl, V.~Satish, M.~Danielczuk, B.~DeRose, S.~McKinley, and
  K.~Goldberg, ``Learning ambidextrous robot grasping policies,'' \emph{Science
  Robotics}, vol.~4, no.~26, 2019.

\bibitem{park2020real}
D.~Park, Y.~Seo, and S.~Y. Chun, ``Real-time, highly accurate robotic grasp
  detection using fully convolutional neural network with rotation ensemble
  module,'' in \emph{2020 IEEE International Conference on Robotics and
  Automation (ICRA)}.\hskip 1em plus 0.5em minus 0.4em\relax IEEE, 2020, pp.
  9397--9403.

\bibitem{cheng2020high}
H.~Cheng, D.~Ho, and M.~Q.-H. Meng, ``High accuracy and efficiency grasp pose
  detection scheme with dense predictions,'' in \emph{2020 IEEE International
  Conference on Robotics and Automation (ICRA)}.\hskip 1em plus 0.5em minus
  0.4em\relax IEEE, 2020, pp. 3604--3610.

\bibitem{baichuanvisual2022}
\BIBentryALTinterwordspacing
B.~Huang, S.~D. Han, J.~Yu, and A.~Boularias, ``Visual foresight trees for
  object retrieval from clutter with nonprehensile rearrangement,'' \emph{IEEE
  Robotics and Automation Letters}, vol.~7, no.~1, p. 231–238, Jan 2022.
  [Online]. Available: \url{http://dx.doi.org/10.1109/lra.2021.3123373}
\BIBentrySTDinterwordspacing

\bibitem{huangdipn2021}
\BIBentryALTinterwordspacing
B.~Huang, S.~D. Han, A.~Boularias, and J.~Yu, ``Dipn: Deep interaction
  prediction network with application to clutter removal,'' \emph{2021 IEEE
  International Conference on Robotics and Automation (ICRA)}, May 2021.
  [Online]. Available: \url{http://dx.doi.org/10.1109/ICRA48506.2021.9561073}
\BIBentrySTDinterwordspacing

\bibitem{huang2022selfsupervised}
B.~Huang, T.~Guo, A.~Boularias, and J.~Yu, ``Self-supervised monte carlo tree
  search learning for object retrieval in clutter,'' \emph{arXiv}, 2022.

\bibitem{rodriguez2018transferring}
D.~Rodriguez and S.~Behnke, ``Transferring category-based functional grasping
  skills by latent space non-rigid registration,'' \emph{IEEE Robotics and
  Automation Letters}, vol.~3, no.~3, pp. 2662--2669, 2018.

\bibitem{liang2019pointnetgpd}
H.~Liang, X.~Ma, S.~Li, M.~G{\"o}rner, S.~Tang, B.~Fang, F.~Sun, and J.~Zhang,
  ``{Point}{Net}{GPD}: Detecting grasp configurations from point sets,'' in
  \emph{IEEE International Conference on Robotics and Automation (ICRA)}, 2019.

\bibitem{mousavian20196}
A.~Mousavian, C.~Eppner, and D.~Fox, ``6-dof graspnet: Variational grasp
  generation for object manipulation,'' in \emph{Proceedings of the IEEE/CVF
  International Conference on Computer Vision}, 2019, pp. 2901--2910.

\bibitem{qin2020s4g}
Y.~Qin, R.~Chen, H.~Zhu, M.~Song, J.~Xu, and H.~Su, ``S4g: Amodal single-view
  single-shot se (3) grasp detection in cluttered scenes,'' in \emph{Conference
  on robot learning}.\hskip 1em plus 0.5em minus 0.4em\relax PMLR, 2020, pp.
  53--65.

\bibitem{fang2020graspnet}
H.-S. Fang, C.~Wang, M.~Gou, and C.~Lu, ``Graspnet-1billion: A large-scale
  benchmark for general object grasping,'' in \emph{Proceedings of the IEEE/CVF
  conference on computer vision and pattern recognition}, 2020, pp.
  11\,444--11\,453.

\bibitem{breyer2020volumetric}
M.~Breyer, J.~J. Chung, L.~Ott, S.~Roland, and N.~Juan, ``Volumetric grasping
  network: Real-time 6 dof grasp detection in clutter,'' in \emph{Conference on
  Robot Learning}, 2020.

\bibitem{sundermeyer2021contact}
M.~Sundermeyer, A.~Mousavian, R.~Triebel, and D.~Fox, ``Contact-graspnet:
  Efficient 6-dof grasp generation in cluttered scenes,'' \emph{2021 IEEE
  International Conference on Robotics and Automation (ICRA)}, 2021.

\bibitem{do2018affordancenet}
T.-T. Do, A.~Nguyen, and I.~Reid, ``Affordancenet: An end-to-end deep learning
  approach for object affordance detection,'' in \emph{2018 IEEE international
  conference on robotics and automation (ICRA)}.\hskip 1em plus 0.5em minus
  0.4em\relax IEEE, 2018, pp. 5882--5889.

\bibitem{detry2017task}
R.~Detry, J.~Papon, and L.~Matthies, ``Task-oriented grasping with semantic and
  geometric scene understanding,'' in \emph{2017 IEEE/RSJ International
  Conference on Intelligent Robots and Systems (IROS)}.\hskip 1em plus 0.5em
  minus 0.4em\relax IEEE, 2017, pp. 3266--3273.

\bibitem{antanas2019semantic}
L.~Antanas, P.~Moreno, M.~Neumann, R.~P. de~Figueiredo, K.~Kersting,
  J.~Santos-Victor, and L.~De~Raedt, ``Semantic and geometric reasoning for
  robotic grasping: a probabilistic logic approach,'' \emph{Autonomous Robots},
  vol.~43, no.~6, pp. 1393--1418, 2019.

\bibitem{kokic2017affordance}
M.~Kokic, J.~A. Stork, J.~A. Haustein, and D.~Kragic, ``Affordance detection
  for task-specific grasping using deep learning,'' in \emph{2017 IEEE-RAS 17th
  International Conference on Humanoid Robotics (Humanoids)}.\hskip 1em plus
  0.5em minus 0.4em\relax IEEE, 2017, pp. 91--98.

\bibitem{ardon2020self}
P.~Ard{\'o}n, E.~Pairet, Y.~Petillot, R.~P. Petrick, S.~Ramamoorthy, and K.~S.
  Lohan, ``Self-assessment of grasp affordance transfer,'' in \emph{2020
  IEEE/RSJ International Conference on Intelligent Robots and Systems
  (IROS)}.\hskip 1em plus 0.5em minus 0.4em\relax IEEE, 2020, pp. 9385--9392.

\bibitem{xu2021affordance}
R.~Xu, F.-J. Chu, C.~Tang, W.~Liu, and P.~A. Vela, ``An affordance keypoint
  detection network for robot manipulation,'' \emph{IEEE Robotics and
  Automation Letters}, vol.~6, no.~2, pp. 2870--2877, 2021.

\bibitem{kokic2020learning}
M.~Kokic, D.~Kragic, and J.~Bohg, ``Learning task-oriented grasping from human
  activity datasets,'' \emph{IEEE Robotics and Automation Letters}, vol.~5,
  no.~2, pp. 3352--3359, 2020.

\bibitem{allevato2020learning}
A.~Allevato, M.~Pryor, E.~S. Short, and A.~L. Thomaz, ``Learning labeled robot
  affordance models using simulations and crowdsourcing,'' in \emph{Robotics:
  Science and Systems (RSS)}, 2020.

\bibitem{yang2019task}
C.~Yang, X.~Lan, H.~Zhang, and N.~Zheng, ``Task-oriented grasping in object
  stacking scenes with crf-based semantic model,'' in \emph{2019 IEEE/RSJ
  International Conference on Intelligent Robots and Systems (IROS)}.\hskip 1em
  plus 0.5em minus 0.4em\relax IEEE, 2019, pp. 6427--6434.

\bibitem{turpin2021gift}
D.~Turpin, L.~Wang, S.~Tsogkas, S.~Dickinson, and A.~Garg, ``Gift:
  Generalizable interaction-aware functional tool affordances without labels,''
  \emph{RSS}, 2021.

\bibitem{zhao2020towards}
J.~Zhao, D.~Troniak, and O.~Kroemer, ``Towards robotic assembly by predicting
  robust, precise and task-oriented grasps,'' \emph{CoRL}, 2020.

\bibitem{florence2018dense}
P.~R. Florence, L.~Manuelli, and R.~Tedrake, ``Dense object nets: Learning
  dense visual object descriptors by and for robotic manipulation,''
  \emph{CoRL}, 2018.

\bibitem{yang2021learning}
S.~Yang, W.~Zhang, R.~Song, J.~Cheng, and Y.~Li, ``Learning multi-object dense
  descriptor for autonomous goal-conditioned grasping,'' \emph{IEEE Robotics
  and Automation Letters}, vol.~6, no.~2, pp. 4109--4116, 2021.

\bibitem{chai2019multi}
C.-Y. Chai, K.-F. Hsu, and S.-L. Tsao, ``Multi-step pick-and-place tasks using
  object-centric dense correspondences,'' in \emph{2019 IEEE/RSJ International
  Conference on Intelligent Robots and Systems (IROS)}.\hskip 1em plus 0.5em
  minus 0.4em\relax IEEE, 2019, pp. 4004--4011.

\bibitem{pan2012fcl}
J.~Pan, S.~Chitta, and D.~Manocha, ``Fcl: A general purpose library for
  collision and proximity queries,'' in \emph{2012 IEEE International
  Conference on Robotics and Automation}.\hskip 1em plus 0.5em minus
  0.4em\relax IEEE, 2012, pp. 3859--3866.

\bibitem{qi2017pointnet}
C.~R. Qi, H.~Su, K.~Mo, and L.~J. Guibas, ``{PointNet}: Deep learning on point
  sets for 3d classification and segmentation,'' in \emph{Proceedings of the
  IEEE conference on computer vision and pattern recognition}, 2017, pp.
  652--660.

\bibitem{fischler1981random}
M.~A. Fischler and R.~C. Bolles, ``Random sample consensus: a paradigm for
  model fitting with applications to image analysis and automated
  cartography,'' \emph{Communications of the ACM}, vol.~24, no.~6, pp.
  381--395, 1981.

\bibitem{patten2020dgcm}
T.~Patten, K.~Park, and M.~Vincze, ``Dgcm-net: dense geometrical correspondence
  matching network for incremental experience-based robotic grasping,''
  \emph{Frontiers in Robotics and AI}, vol.~7, 2020.

\bibitem{jiang2020pointgroup}
L.~Jiang, H.~Zhao, S.~Shi, S.~Liu, C.-W. Fu, and J.~Jia, ``{PointGroup}:
  Dual-set point grouping for 3d instance segmentation,'' \emph{Proceedings of
  the IEEE Conference on Computer Vision and Pattern Recognition (CVPR)}, 2020.

\bibitem{SubmanifoldSparseConvNet}
B.~Graham and L.~van~der Maaten, ``Submanifold sparse convolutional networks,''
  \emph{arXiv preprint arXiv:1706.01307}, 2017.

\bibitem{xie2021unseen}
C.~Xie, Y.~Xiang, A.~Mousavian, and D.~Fox, ``Unseen object instance
  segmentation for robotic environments,'' \emph{IEEE Transactions on
  Robotics}, 2021.

\bibitem{ester1996density}
M.~Ester, H.-P. Kriegel, J.~Sander, X.~Xu \emph{et~al.}, ``A density-based
  algorithm for discovering clusters in large spatial databases with noise.''
  in \emph{Kdd}, vol.~96, no.~34, 1996, pp. 226--231.

\bibitem{coumans2021}
E.~Coumans and Y.~Bai, ``Pybullet, a python module for physics simulation for
  games, robotics and machine learning,'' \url{http://pybullet.org},
  2016--2021.

\bibitem{tremblay2018deep}
J.~Tremblay, T.~To, B.~Sundaralingam, Y.~Xiang, D.~Fox, and S.~Birchfield,
  ``Deep object pose estimation for semantic robotic grasping of household
  objects,'' \emph{CoRL}, 2018.

\bibitem{ha2020fit2form}
H.~Ha, S.~Agrawal, and S.~Song, ``{Fit2Form}: 3{D} generative model for robot
  gripper form design,'' in \emph{Conference on Robotic Learning (CoRL)}, 2020.

\bibitem{antonova2018global}
R.~Antonova, M.~Kokic, J.~A. Stork, and D.~Kragic, ``Global search with
  bernoulli alternation kernel for task-oriented grasping informed by
  simulation,'' \emph{CoRL}, 2018.

\bibitem{wen2021bundletrack}
B.~Wen and K.~Bekris, ``Bundletrack: 6d pose tracking for novel objects without
  instance or category-level 3d models,'' \emph{IROS}, 2021.

\end{thebibliography}

\end{document}